\documentclass[a4paper,11pt]{article}

\usepackage[colorlinks = true,
            linkcolor = magenta,
            urlcolor  = blue,
            citecolor = green,
            anchorcolor = blue]{hyperref}

\usepackage{amsmath,amssymb,amsthm}
\usepackage{authblk}  % authors with affiliations
\usepackage[british]{babel}
\usepackage{bbm}  % bold numbers in math mode, e.g. for the indicator function
\usepackage[sorting=none,doi=false,isbn=false,giveninits=true,style=nature]{biblatex}           % modern bibliography
\usepackage{caption}
\usepackage{color}
\usepackage{enumitem}
\usepackage[T1]{fontenc}
\usepackage[top=2.5cm, bottom=2.5cm, left=2.5cm, right=2.5cm]{geometry}
\usepackage{graphicx}
\usepackage{graphbox}       % easy vertical alignment of graphics
\PassOptionsToPackage{hyphens}{url}\usepackage{hyperref}
\usepackage[utf8]{inputenc}
\usepackage{csquotes}  % needs to be loaded *after* inputenc
\usepackage[mono=false]{libertine}
\usepackage{nicefrac}
\usepackage{mathtools}
\usepackage{microtype}      % microtypography
\usepackage{subcaption}
\usepackage{xcolor}

\usepackage{multirow}
\usepackage{array, booktabs, makecell}

\bibliography{refs}

\newcommand{\E}{\mathbb{E}}

\newcommand{\R}{\mathbb{R}}

\newcommand{\bs}{\boldsymbol}
\newcommand{\citet}[1]{\textcite{#1}}

\title{Transformed CNNs: \\recasting pre-trained convolutional layers with self-attention}

\author[1,2]{St\'ephane d'Ascoli\thanks{stephane.dascoli@ens.fr}}
\author[2]{Levent Sagun}
\author[1]{Giulio Biroli}
\author[2]{Ari Morcos}
\affil[1]{Laboratoire de Physique de l’Ecole Normale Supérieure, Université
  PSL, CNRS, \protect \\ Sorbonne Université, Université Paris-Diderot, Sorbonne Paris
  Cité, Paris, France}
\affil[2]{Facebook AI Research, Paris, France}
\date{}

\begin{document}

\maketitle
\begin{abstract}
Vision Transformers (ViT) have recently emerged as a powerful alternative to convolutional networks (CNNs). Although hybrid models attempt to bridge the gap between these two architectures, the self-attention layers they rely on induce a strong computational bottleneck, especially at large spatial resolutions. In this work, we explore the idea of reducing the time spent training these layers by initializing them as convolutional layers. This enables us to transition smoothly from any pre-trained CNN to its functionally identical hybrid model, called Transformed CNN (T-CNN). With only 50 epochs of fine-tuning, the resulting T-CNNs demonstrate significant performance gains over the CNN (+2.2\% top-1 on ImageNet-1k for a ResNet50-RS) as well as substantially improved robustness (+11\% top-1 on ImageNet-C). We analyze the representations learnt by the T-CNN, providing deeper insights into the fruitful interplay between convolutions and self-attention. Finally, we experiment initializing the T-CNN from a partially trained CNN, and find that it reaches better performance than the corresponding hybrid model trained from scratch, while reducing training time.
\end{abstract}

\section*{Introduction}

Since the success of AlexNet in 2012~\cite{krizhevsky2017imagenet}, the field of Computer Vision has been dominated by Convolutional Neural Networks (CNNs)~\cite{lecun1998gradient,lecun1989backpropagation}. Their local receptive fields give them a strong inductive bias to exploit the spatial structure of natural images~\cite{scherer_evaluation_2010,schmidhuber_deep_2015,goodfellow_deep_2016}, while allowing them to scale to large resolutions seamlessly. Yet, this inductive bias limits their ability to capture long-range interactions. 

In this regard, self-attention (SA) layers, originally introduced in language models~\cite{bahdanau2014neural,vaswani2017attention,devlin2018bert}, have gained interest as a building block for vision~\citet{ramachandran2019stand,zhao2020exploring}. Recently, they gave rise to a plethora of Vision Transformer (ViT) models, able to compete with state-of-the-art CNNs in various tasks~\citet{dosovitskiy2020image,touvron2020training,wu_visual_2020,touvron2021going,liu2021swin,heo2021rethinking} while demonstrating better robustness~\cite{bhojanapalli2021understanding,mao2021rethinking}. However, capturing long-range dependencies necessarily comes at the cost of quadratic complexity in input size, a computational burden which many recent directions have tried to alleviate~\cite{bello2021lambdanetworks,wang2020self,choromanski2020rethinking,katharopoulos2020transformers}. Additionally, ViTs are generally harder to train~\cite{zhang2019adaptive,liu2020understanding}, and require vast amounts of pre-training~\cite{dosovitskiy2020image} or distillation from a convolutional teacher~\cite{hinton2015distilling,jiang2021token,graham2021levit} to match the performance of CNNs.

Faced with the dilemma between efficient CNNs and powerful ViTs, several approaches have aimed to bridge the gap between these architectures. On one side, hybrid models append SA layers onto convolutional backbones ~\cite{chen20182,bello2019attention,graham2021levit,chen2021visformer,srinivas2021bottleneck}, and have already fueled successful results in a variety of tasks~\cite{carion2020end,hu2018relation,chen2020uniter,locatello2020object,sun2019videobert}. Conversely, a line of research has studied the benefit of introducing convolutional biases in Transformer architectures to ease learning~\cite{d2021convit,wu2021cvt,yuan2021incorporating}.
Despite these interesting compromises, modelling long-range dependencies at low computational cost remains a challenge for practitioners.

\begin{figure}
    \centering
    \includegraphics[width=\linewidth]{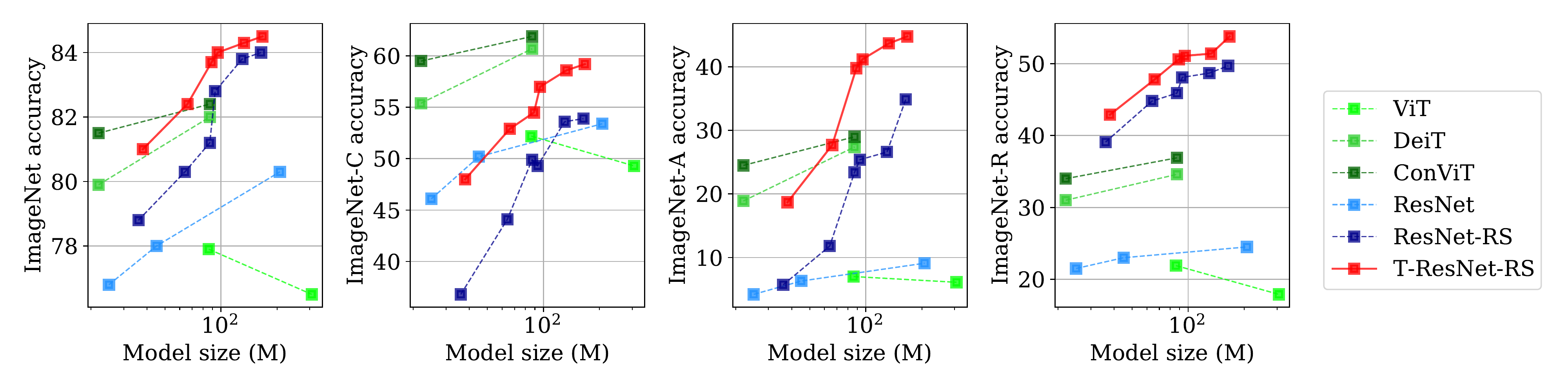}
    \caption{\textbf{Transformed ResNets strike a strong accuracy-robustness balance.} Our models (red) significantly outperform the original ResNet-RS models (dark blue) they were initialized from when evaluated on ImageNet-1k. On various robustness benchmarks (ImageNet-C, A and R, from left to right), they narrow or close the gap with Transformer architectures.}
    \label{fig:intro}
\end{figure}

\paragraph{Contributions}
At a time when pre-training on vast datasets has become common practice, we ask the following question: does one need to train the SA layers during the whole learning process? Could one instead learn cheap components such as convolutions first, leaving the SA layers to be learnt at the end? In this paper, we take a step in this direction by presenting a method to fully reparameterize a pre-trained convolutional layer as a \textit{Gated Positional Self-Attention} (GPSA) layer~\cite{d2021convit}. The latter is initialized to reproduce the mapping of the convolutional layer, but is then encouraged to learn more general mappings which are not accessible to the CNN by adjusting positional gating parameters.

We leverage this method to reparametrize pre-trained CNNs as functionally equivalent hybrid models. After only 50 epochs of fine-tuning, the resulting Transformed CNNs (T-CNNs) boast significant performance and robustness improvements as shown in Fig.~\ref{fig:intro}, demonstrating the practical relevance of our method. We analyze the inner workings of the T-CNNs, showing how they learn more robust representations by combining convolutional heads and SA heads in a complementary way. Finally, we investigate how performance gains depend on the reparametrization epoch. Results suggest that reparametrizing at intermediate times is optimal in terms of speed-performance trade-offs.

% Our code is included in the SM.

\paragraph{Related work}

Our work mainly builds on two pillars. First, the idea that SA layers can express any convolution, introduced by~\citet{cordonnier2019relationship}.  This idea was recently leveraged in~\citet{d2021convit}, which initialize the SA layers of the ViT as \textit{random} convolutions and observe performance gains compared to the standard initialization, especially in the low-data regime where inductive biases are most useful. Our approach is a natural follow-up of this idea: what happens if the SA layers are instead initialized as \textit{trained} convolutions?

Second, we exploit the following learning paradigm: train a simple and fast model, then reparameterize it as a more complex model for the final stages of learning. This approach was studied from a scientific point of view in~\citet{d2019finding}, which shows that reparameterizing a CNN as a fully-connected network (FCN) halfway through training can lead the FCN to outperform the CNN. Yet, the practical relevance of this method is limited by the vast increase in number of parameters required by the FCN to functionally represent the CNN. In contrast, our reparameterization hardly increases the parameter count of the CNN, making it easily applicable to any state-of-the-art CNN. Note that these reparameterization methods can be viewed an informed version of dynamic architecture growing algorithms such as AutoGrow~\cite{wen2020autogrow}.

In the context of hybrid models, various works have studied the performance gains obtained by introducing MHSA layers in ResNets with minimal architectural changes~\cite{srinivas2021bottleneck,graham2021levit,chen2021visformer}. However, the MHSA layers used in these works are initialized randomly and need to be trained from scratch. Our approach is different, as it makes use of GPSA layers, which can be initialized to represent the same function as the convolutional layer it replaces. We emphasize that the novelty in our work is not in the architectures used, but in the unusual way they are blended together. %\ari{Should we include bottleneck transformer or related as a baseline? SD: Don't have the performance, as model isn't in pytorch package} 

\vspace{-2mm}
\section{Background}

% We begin by a reminder on SA layers and how positional attention can allow them to express convolutional layers. 

\paragraph{Multi-head self-attention} \label{sec:mhsa}

The SA mechanism is based on a trainable associative memory with (key, query) vector pairs. To extract the semantic interpendencies between the $L$ elements of a sequence $\bs X\in\R^{L\times D_{in}}$, a sequence of ``query'' embeddings $\boldsymbol Q = \boldsymbol W_{qry} \boldsymbol X\in \mathbb{R}^{L\times D_{h}}$ is matched against another sequence of ``key'' embeddings $\boldsymbol K = \boldsymbol W_{key} \boldsymbol X\in\mathbb{R}^{L\times D_{h}}$ using inner products. The result is an attention matrix whose entry $(ij)$ quantifies how semantically relevant $\boldsymbol Q_i$ is to $\boldsymbol K_j$:
\begin{equation}
    \boldsymbol A = \operatorname{softmax}\left(\frac{\boldsymbol Q \boldsymbol K^\top}{\sqrt {D_{h}}}\right) \in \mathbb{R
}^{L\times L}.
    \label{eq:attention}
\end{equation}

Multi-head SA layers use several SA heads in parallel to allow the learning of different kinds of dependencies:
\begin{align}
    \operatorname{MSA}(\boldsymbol{X}):=\sum_{h=1}^{N_h} \left[\text{SA}_{h}(\boldsymbol{X})\right] \boldsymbol{W}^h_{out}, \quad\quad  \text{SA}_h(\boldsymbol{X}) := \boldsymbol{A}^h \boldsymbol{X} \boldsymbol{W}_{val}^h,
\end{align}
where $\bs W_\text{val}^h \in R^{D_{in}\times D_{v}}$ and $\bs W_\text{out}^h \in R^{D_{v}\times D_{out}}$ are two learnable projections. 

To incorporate positional information, ViTs usually add absolute position information to the input at embedding time, before propagating it through the SA layers. Another possibility is to replace the vanilla SA with positional SA (PSA), including a position-dependent term in the softmax~\cite{ramachandran2019stand, shaw2018self}. Although there are several way to parametrize the positional attention, we use encodings $\boldsymbol r_{ij}$ of the relative position of pixels $i$ and $j$ as in~\cite{cordonnier2019relationship,d2021convit}:
\begin{align}
    \boldsymbol{A}^h_{ij}:=\operatorname{softmax}\left(\boldsymbol Q^h_i \boldsymbol K^{h\top}_j+\boldsymbol{v}_{pos}^{h\top} \boldsymbol{r}_{ij}\right).
\label{eq:local-attention}
\end{align}
Each attention head learns an embedding $\boldsymbol{v}_{pos}^h \in \mathbb R^{D_{pos}}$, and the relative positional encodings $\boldsymbol{r}_{ij}\in \mathbb R^{D_{pos}}$ only depend on the distance between pixels $i$ and $j$,  denoted denoted as a two-dimensional vector $\boldsymbol \delta_{ij}$. 

\paragraph{Self-attention as a generalized convolution}

\citet{cordonnier2019relationship} shows that a multi-head PSA layer (Eq. \ref{eq:local-attention}) with $N_h$ heads and dimension $D_{pos}\geq 3$ can express any convolutional layer of filter size $\sqrt N_h$, with $D_{in}$ input channels and $\min(D_v, D_{out})$ output channels, by setting the following:
\begin{align}
    \begin{cases}
    &\boldsymbol{v}_{pos}^{h}:=-\alpha^{h}\left(1,-2 \Delta_{1}^{h},-2 \Delta_{2}^{h}, 0,\ldots 0\right)\\ &\boldsymbol{r}_{\boldsymbol{\delta}}:=\left(\|\boldsymbol{\delta}\|^{2}, \delta_{1}, \delta_{2},0,\ldots 0\right)\\
    &\boldsymbol{W}_{q r y}=\boldsymbol{W}_{k e y}:=\mathbf{0}
    \end{cases}
    \label{eq:local-init}
\end{align}

In the above, the \emph{center of attention} $\boldsymbol \Delta^h\in\mathbb{R}^2$ is the position to which head $h$ pays most attention to, relative to the query pixel, whereas the \emph{locality strength} $\alpha^h>0$ determines how focused the attention is around its center $\boldsymbol \Delta^h$. When $\alpha^h$ is large, the attention is focused only on the pixel located at $\boldsymbol \Delta^h$; when $\alpha^h$ is small, the attention is spread out into a larger area. Thus, the PSA layer can achieve a convolutional attention map by setting the centers of attention $\boldsymbol \Delta^h$ to each of the possible positional offsets of a $\sqrt{N_h}\times \sqrt{N_h}$ convolutional kernel, and sending the locality strengths $\alpha^h$ to some large value.

\section{Approach}

In this section, we introduce our method for mapping a convolutional layer to a functionally equivalent PSA layer with minimal increase in parameter count. To do this, we leverage the GPSA layers introduced in~\citet{d2021convit}. 

\paragraph{Loading the filters}

We want each head $h$ of the PSA layer to functionally mimic the pixel $h$ of a convolutional filter $\bs W_\text{filter} \in \R^{N_h \times D_{in} \times D_{out}}$, where we typically have $D_{out}\geq D_{in}$. Rewriting the action of the MHSA operator in a more explicit form, we have
\begin{equation}
    \operatorname{MHSA}(\boldsymbol{X})=\sum_{h =1}^{N_h} \boldsymbol{A}^{h} \boldsymbol{X} \underbrace{\boldsymbol{W}_{\text {val }}^{h} \boldsymbol{W}_{\text {out }}^h}_{\boldsymbol{W}^{h}\in\R^{D_{in}\times D_{out}}}
\end{equation}
In the convolutional configuration of Eq.~\ref{eq:local-init}, $\bs A^h \bs X$ selects pixel $h$ of $\bs X$. Hence, we need to set $\bs W^h=\bs W_\text{filter}^h$. However, as a product of matrices, the rank of $\bs W_h$ is bottlenecked by $D_v$. To avoid this being a limitation, we need $D_v \geq D_{in}$ (since $D_{out}\geq D_{in}$). To achieve this with a minimal number of parameters, we choose $D_v = D_{in}$, and simply set the following initialization:
\begin{align}
    \bs W_\text{val}^h = \bs I, \quad\quad\quad \bs W_\text{out}^h = \bs W_\text{filter}^h.
\end{align}
Note that this differs from the usual choice made in SA layers, where $D_v = \lfloor D_{in}/N_h \rfloor$. However, to keep the parameter count the same, we share the same $\bs W_{val}^h$ across different heads $h$, since it plays a symmetric role at initialization.

Note that this reparameterization introduces three additional matrices compared to the convolutional filter: $\bs W_{qry}, \bs W_{key}, \bs W_{val}$, each containing $ D_{in} \times D_{in}$ parameters. However, since the convolutional filter contains $N_h \times D_{in} \times D_{out}$ parameters, where we typically have $N_h=9$ and $D_{out}\in\{D_{in},2D_{in}\}$, these additional matrices are much smaller than the filters and hardly increase the parameter count. This can be seen from the model sizes in Tab.~\ref{tab:robustness}.

\paragraph{Gated Positional self-attention}
Recent work~\cite{d2021convit} has highlighted an issue with standard PSA: the fact that the content and positional terms in Eq.~\ref{eq:local-attention} are potentially of very different magnitudes, in which case the softmax ignores the smallest of the two. This can typically lead the PSA to adopt a greedy attitude: choosing the form of attention (content or positional) which is easiest at a given time then sticking to it.

To avoid this, the ConViT~\citet{d2021convit} uses GPSA layers which sum the content and positional terms \emph{after} the softmax, with their relative importances governed by a learnable \emph{gating} parameter $\lambda_h$ (one for each attention head). In GPSA layers, the attention is parametrized as follows:
\begin{align}
    % \text{GPSA}_h(\boldsymbol{X}) :=& \operatorname{normalize}\left[\boldsymbol{A}^h\right] \boldsymbol{X} \boldsymbol{W}_{val}^h\\
    \boldsymbol{A}^h_{ij}:=&\left(1-\sigma(\lambda_h)\right) \operatorname{softmax}\left(\boldsymbol Q^h_i \boldsymbol K^{h\top}_j\right) + \sigma(\lambda_h) \operatorname{softmax}\left(\boldsymbol{v}_{pos}^{h\top} \boldsymbol{r}_{ij}\right),
    \label{eq:gating-param}
\end{align}
where $\sigma:x\mapsto \nicefrac{1}{\left(1+e^{-x}\right)}$ is the sigmoid function. In the positional part, the encodings $\bs r_{ij}$ are fixed rather than learnt (see Eq.~\ref{eq:local-init}), which makes changing input resolution straightforward (see SM.~\ref{app:resolution}) and leaves only 3 learnable parameters per head: $\bs \Delta_1, \bs \Delta_2$ and $\alpha$\footnote{Since $\alpha$ represents the temperature of the softmax, its value must stay positive at all times. To ensure this, we instead learn a rectified parameter $\tilde \alpha$ using the softplus function: $\alpha = \frac{1}{\beta} \log (1+e^{-\beta \tilde \alpha})$, with $\beta=5$.}. 

\paragraph{How convolutional should the initialization be?}

The convolutional initialization of GPSA layers involves two parameters, determining how strictly convolutional the behavior is: the initial value of the \emph{locality strength} $\alpha$, which determines how focused each attention head is on its dedicated pixel, and the initial value of the \emph{gating parameters} $\lambda$, which determines the importance of the positional information versus content. If $\lambda_h\gg 0$ and $\alpha\gg 1$, the T-CNN will perfectly reproduce the input-output function of the CNN, but may stay stuck in the convolutional configuration. Conversely, if $\lambda_h\ll 0$ and $\alpha\ll 1$, the T-CNN will poorly reproduce the input-output function of the CNN. Hence, we choose $\alpha =1$ and $\lambda = 1$ to lie in between these two extremes. This puts the T-CNN ``on the verge of locality'', enabling it to escape locality effectively throughout training.

\paragraph{Architectural details}
To make our setup as canonical as possible, we focus on ResNet architectures~\cite{he2016deep}, which contain 5 stages, with spatial resolution halfed and number of channels doubled at each stage. Our method involves reparameterizing $3\times 3$ convolutions as GPSA layers with 9 attention heads. However, global SA is too costly in the first layers, where the spatial resolution is large. We therefore only reparameterize the last stage of the architecture, while replacing the first stride-2 convolution by a stride-1 convolution, exactly as in~\cite{srinivas2021bottleneck}. We also add explicit padding layers to account for the padding of the original convolutions.

% \paragraph{How far should we escape locality?}

% Once we have chosen the correct $\alpha$ and $\lambda$ for the initialization, we need to adjust how far the T-CNN escapes the CNN function space. This can be done by adjusting a specific learning rate of for the gating parameters $\lambda_h$. Again, a balance needs to be found between lazily staying in the convolutional configuration and catastrophically forgetting it. Empirically, we observed that the standard learning rate used for the fine-tuning $\eta = 0.0001$ was too low to let the gating parameters escape locality. We found the optimal learning rate to do so to be $\eta_\lambda = 0.1$.

\section{Performance of the Transformed CNNs}
\label{sec:finetuning}

In this section, we apply our reparametrization to state-of-the-art CNNs, then fine-tune the resulting T-CNNs to learn better representations. This method allows to fully disentangle the training of the SA layers from that of the convolutional backbone, which is of practical interest for two reasons. First, it minimizes the time spent training the SA layers, which typically have a slower throughput. Second, it separates the algorithmic choices of the CNN backbone from those of the SA layers, which are typically different; for example, CNNs are typically trained with SGD whereas SA layers perform much better with adaptive optimizers such as Adam~\cite{zhang2019adaptive}, an incompatibility which may limit the performance of usual hybrid models. 

\paragraph{Training details}
To minimize computational cost, we restrict the fine-tuning to 50 epochs\footnote{We study how performance depends on the number of fine-tuning epochs in SM.~\ref{app:epochs}.}. Following~\cite{zhang2019adaptive}, we use the AdamW optimizer, with a batch size of 1024\footnote{Confirming the results of~\cite{zhang2019adaptive}, we obtained worse results with SGD.}. The learning rate is warmed up to $10^{-4}$ then annealed using a cosine decay. To encourage the T-CNN to escape the convolutional configuration and learn content-based attention, we use a larger learning rate of 0.1 for the gating parameters of Eq.~\ref{eq:gating-param} (one could equivalently decrease the temperature of the sigmoid function). 

We use the same data augmentation scheme as the DeiT~\cite{touvron2020training}, as well as rather large stochastic depth coefficients $d_r$ reported in Tab.~\ref{tab:finetune}. Hoping that our method could be used as an alternative to the commonly used practice of fine-tuning models at higher resolution, we also increase the resolution during fine-tuning~\cite{touvron2019fixing}. In this setting, a ResNet50 requires only 6 hours of fine-tuning on 16 V100 GPUs, compared to 33 hours for the original training. For our largest model (ResNet350-RS), the fine-tuning lasts 50 hours.

\paragraph{Performance gains}

\begin{figure}
    \centering
    \begin{subfigure}[b]{.48\textwidth}
        \includegraphics[width=\linewidth]{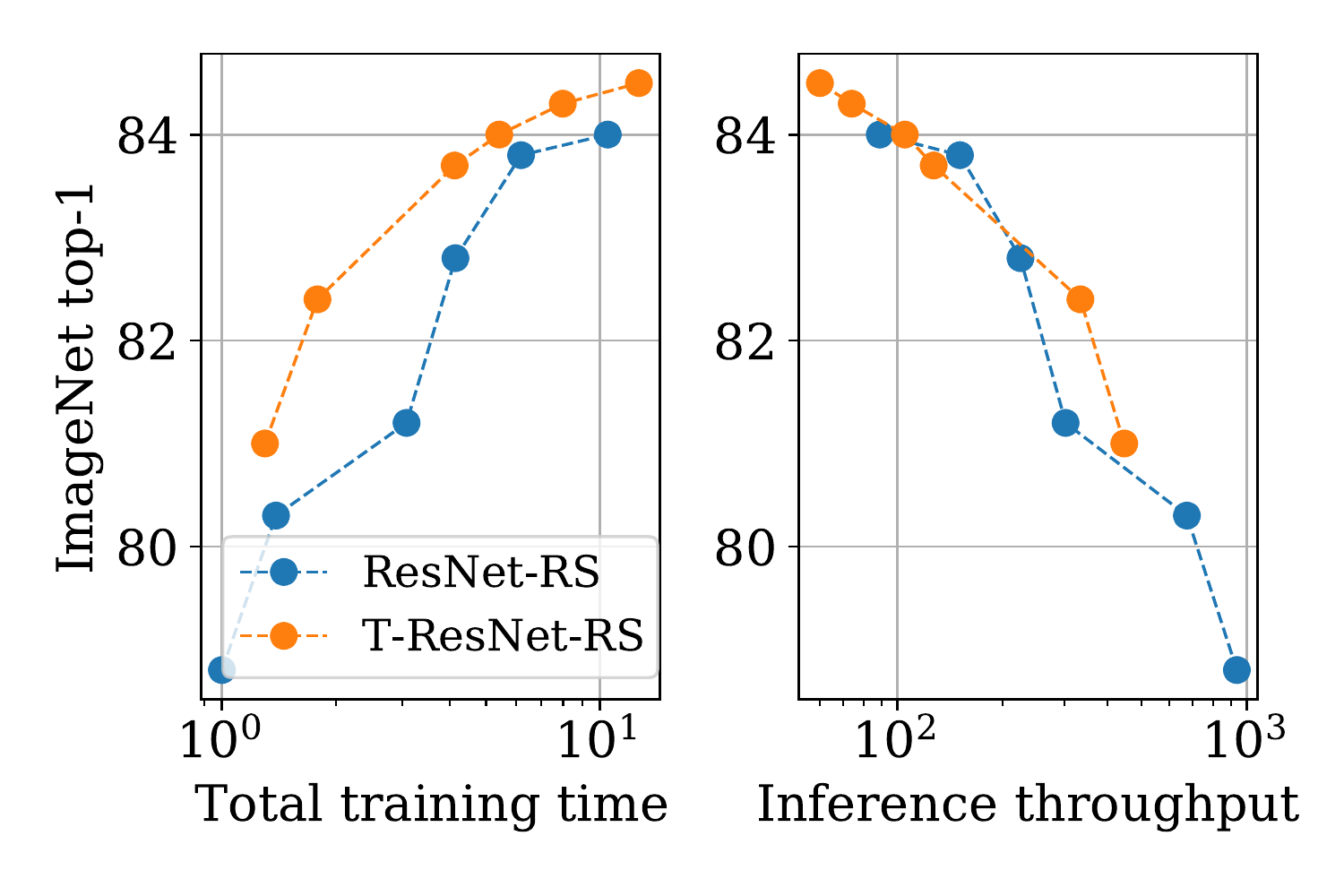}
        %\vspace*{.cm}
        \caption{ImageNet-1k}
    \end{subfigure}
    \begin{subfigure}[b]{.48\textwidth}
        \includegraphics[width=\linewidth]{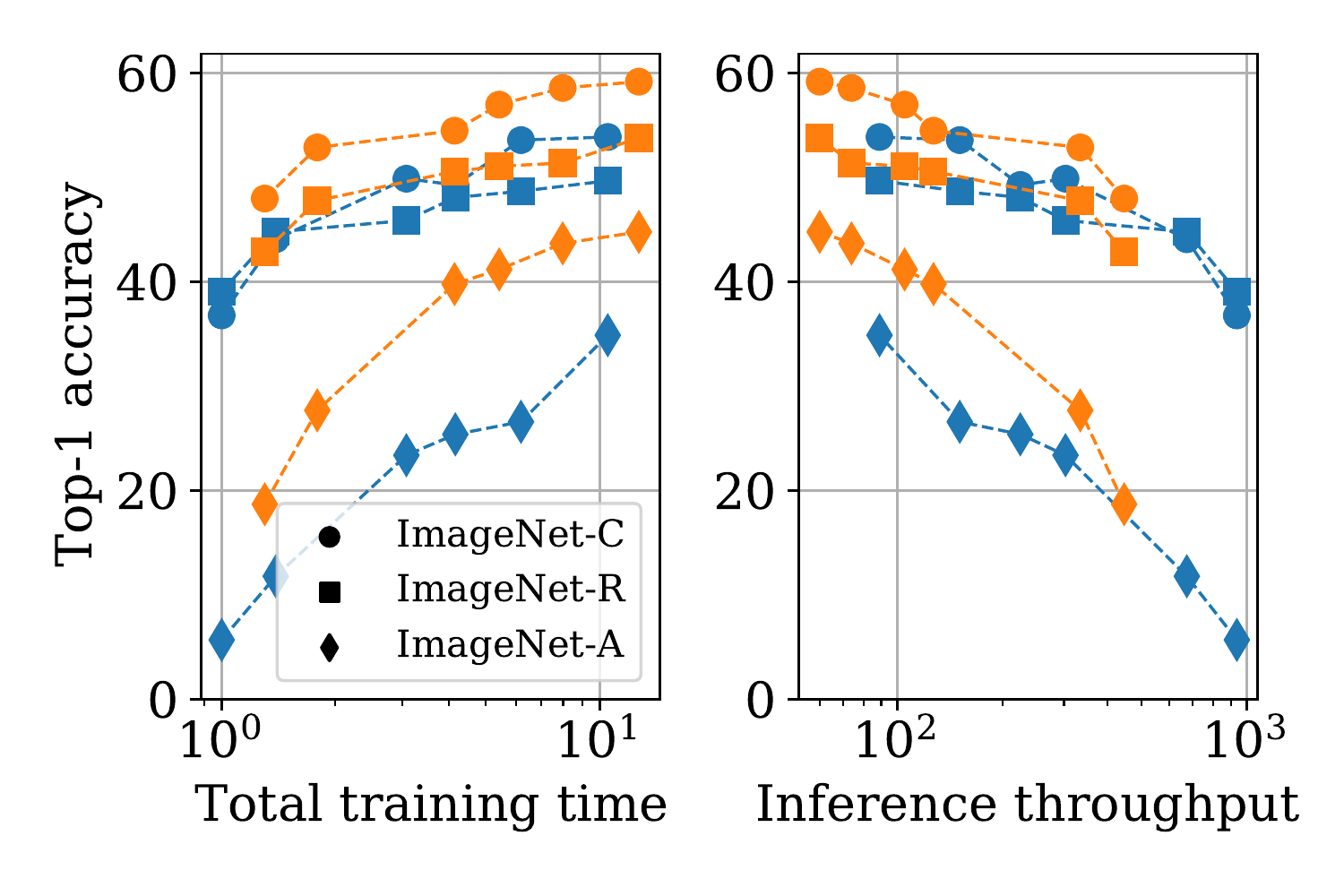}
        %\vspace*{.cm}
        \caption{Robustness benchmarks}
    \end{subfigure}
    \caption{\textbf{T-CNNs present better speed-accuracy trade-offs than the CNNs they stem from.} Total training time (original training + finetuning) is normalized by the total training time of the ResNet50-RS. Inference throughput is the number of images processed per second on a V100 GPU at batch size 32.}
    \label{fig:pareto}
\end{figure}

We applied our method to pre-trained ResNet-RS~\cite{bello2021revisiting} models, using the weights provided by the timm package~\cite{rw2019timm}. These models are derived from the original ResNet~\cite{he2016deep}, but use improved architectural features and training strategies, enabling them to reach better speed-accuracy trade-offs than EfficientNets. Results are presented in Tab.~\ref{tab:finetune}, where we also report the baseline improvement of fine-tuning in the same setting but without SA. In all cases, our fine-tuning improves top-1 accuracy, with a significant gap over the baseline. To demonstrate the wide applicability of our method, we report similar improvements for ResNet-D architectures in SM.~\ref{app:resnetd}.

Despite the extra fine-tuning epochs and their slower throughput, the resulting T-CNNs match the performance of the original CNNs at equal throughput, while significantly outperforming them at equal total training time, as shown in the Pareto curves of Fig.~\ref{fig:pareto}(a)\footnote{We estimated the training times of the original ResNet-RS models based on their throughput, for the same hardware as used for the T-ResNet-RS.}. However, the major benefit of the reparametrization is in terms of robustness, as shown in Fig.~\ref{fig:pareto}(b) and explained below.

\begin{table}[tb]
    %\small
    \centering
    \begin{tabular}{{c|cc|cc|cc|cc|cc}}
    \toprule
    \multirow{3}{*}{\textbf{Backbone}} & \multicolumn{4}{c|}{Training} & \multicolumn{6}{c}{Fine-tuning} \\\cline{2-11}
    \rule{0pt}{3ex}    
    &&&&&&& \multicolumn{2}{c|}{Without SA} & \multicolumn{2}{c}{With SA} \\
    & Res. & $d_r$ & TTT & Top-1 & Res. & $d_r$ & TTT & Top-1 & TTT & Top-1 \\
    \midrule
    ResNet50-RS    & 160 & 0.0 & 1   (ref.) & 78.8 & 224 & 0.1 & 1.16   & 80.4  & 1.30 & \textbf{81.0}\\
    ResNet101-RS   & 192 & 0.0 & 1.39       & 80.3 & 224 & 0.1 & 1.65   & 81.9  & 1.79 & \textbf{82.4}\\
    ResNet152-RS   & 256 & 0.0 & 3.08       & 81.2 & 320 & 0.2 & 3.75   & 83.4  & 4.13 & \textbf{83.7}\\
    ResNet200-RS   & 256 & 0.1 & 4.15       & 82.8 & 320 & 0.2 & 5.04   & 83.7  & 5.42 & \textbf{84.0}\\
    ResNet270-RS   & 256 & 0.1 & 6.19       & 83.8 & 320 & 0.2 & 7.49   & 83.9  & 7.98 & \textbf{84.3}\\
    ResNet350-RS   & 288 & 0.1 & 10.49      & 84.0 & 320 & 0.2 & 12.17  & 84.1  & 12.69& \textbf{84.5}\\
    \bottomrule
    \end{tabular}
    \caption{\textbf{Statistics of the models considered, trained from scratch on ImageNet.} Top-1 accuracy is measured on ImageNet-1k validation set. ``TTT'' stands for total training time (including fine-tuning), normalized by the total training time of the ResNet50-RS. $d_r$ is the stochastic depth coefficient used for the various models.
    }
    \label{tab:finetune}
\end{table}

\begin{figure}
    \centering
    % \begin{subfigure}[b]{.35\textwidth}
    %     \includegraphics[width=\linewidth]{figs/pareto_imnetc.pdf}
    %     \vspace*{.cm}
    %     \caption{Speed-robustness trade-offs}
    % \end{subfigure}
    \begin{subfigure}[b]{\textwidth}
        \includegraphics[width=\linewidth]{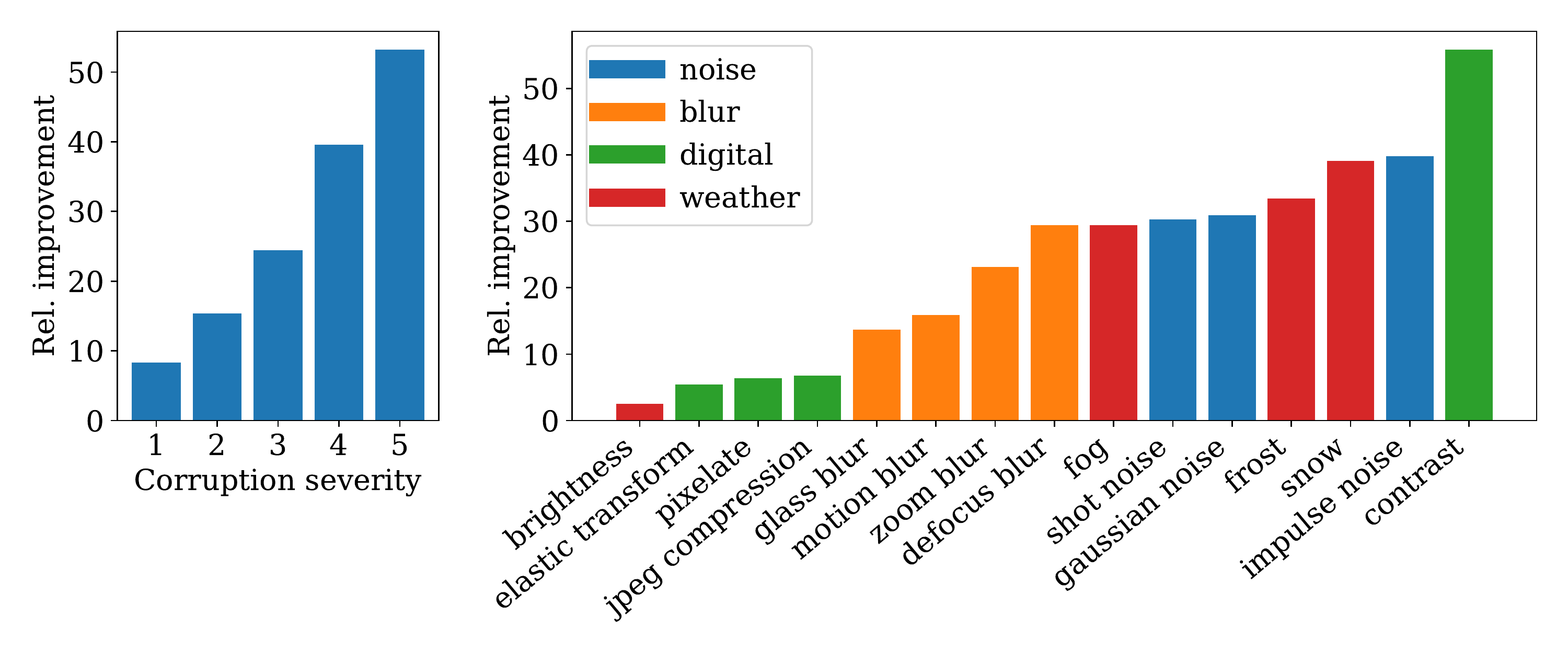}    
    \end{subfigure}
    \caption{\textbf{Robustness is most improved for strong and blurry corruption categories.} We report the relative improvement between the top-1 accuracy of the T-ResNet50-RS and that of the ResNet50-RS on ImageNet-C, averaging over the different corruption categories (left) and corruption severities (right).}
    \label{fig:robustness}
\end{figure}

\newcolumntype{?}{!{\vrule width 1pt}}
\begin{table}
    %\small
    \centering
    \begin{tabular}{c|cccc|cccc}
    \toprule
    \textbf{Model} & Res. & Params & Speed & Flops & ImNet-1k & ImNet-C & ImNet-A & ImNet-R\\\midrule
    \multicolumn{9}{c}{Transformers}\\\midrule
    ViT-B/16       & 224 & 86 M  & 182 & 16.9 & 77.9 & 52.2 & 7.0  & 21.9 \\
    ViT-L/16       & 224 & 307 M & 55  & 59.7 & 76.5 & 49.3 & 6.1  & 17.9 \\\midrule
    DeiT-S         & 224 & 22 M  & 544 & 4.6  & 79.9 & 55.4 & 18.9 & 31.0 \\
    DeiT-B         & 224 & 87 M  & 182 & 17.6 & 82.0 & 60.7 & 27.4 & 34.6 \\\midrule
    ConViT-S       & 224 & 28 M  & 296 & 5.4  & 81.5 & 59.5 & 24.5 & 34.0 \\
    ConViT-B       & 224 & 87 M  & 139 & 17.7 & 82.4 & \textbf{61.9} & 29.0 & 36.9 \\\midrule    % ViT-B/16*      & 86 M  & 182 & 16.9 & 84.0 & 65.8 & 26.7  & 38.0 \\
    % ViT-L/16*      & 307 M & 55  & 59.7 & 85.1 & 70.0 & 28.1  & 40.6 \\\midrule
    % \multicolumn{9}{c}{Reported in~\cite{mao2021rethinking}}\\\midrule
    % RegNetY-4GF    & 20 M  & - & 4.0 & 79.2 & 31.3 & 8.9  & 3.5 &  \\    
    % ResNeXt50-32x4d& 25 M  & - & 4.3 & 79.8 & 35.3 & 10.7 & 3.7\\
    % EfficientNet-B4& 19 M  & - & 4.4 & 83.0 & 28.9 & 26.3 & 4.2\\\midrule
    % DeiT-Ti        & 6 M   & - & 1.3 & 72.2 & 28.9 & 7.3  & 3.3 \\
    % DeiT-S         & 22 M  & - & 4.6 & 79.9 & 45.4 & 18.9  & 3.7 \\
    % DeiT-B         & 87 M. & - & 17.6& 82.0 & 51.5 & 27.4  & 3.8 \\\midrule
    % RvT-Ti         & 11 M  & - & 1.3 & 79.2 & 43.0 & 14.4  & 3.9 \\
    % RvT-S          & 23 M  & - & 4.7 & 81.9 & 50.6 & 25.7  & 4.2 \\
    % RvT-B          & 92 M  & - & 17.7& 82.6 & 53.2 & 28.5  & 4.3 \\\midrule
    \multicolumn{9}{c}{CNNs}\\\midrule
    ResNet50       & 224 & 25 M  & 736 & 4.1  & 76.8 & 46.1 & 4.2  & 21.5 \\
    ResNet101      & 224 & 45 M  & 435 & 7.85 & 78.0 & 50.2 & 6.3  & 23.0 \\
    ResNet101x3    & 224 & 207 M & 62  & 69.6 & 80.3 & 53.4 & 9.1  & 24.5 \\
    ResNet152x4    & 224 & 965 M & 18  & 183.1& 80.4 & 54.5 & 11.6 & 25.8 \\
    \midrule
    ResNet50-RS    & 160 & 36 M  & 938 &  4.6& 78.8 & 36.8 & 5.7  & 39.1 \\
    ResNet101-RS   & 192 & 64 M  & 674 & 12.1& 80.3 & 44.1 & 11.8 & 44.8 \\
    ResNet152-RS   & 256 & 87 M  & 304 & 31.2& 81.2 & 49.9 & 23.4 & 45.9 \\
    ResNet200-RS   & 256 & 93 M  & 225 & 40.4& 82.8 & 49.3 & 25.4 & 48.1 \\
    ResNet270-RS   & 256 & 130 M & 152 & 54.2& 83.8 & 53.6 & 26.6 & 48.7 \\
    ResNet350-RS   & 288 & 164 M &  89 & 87.5& 84.0 & 53.9 & 34.9 & 49.7 \\
    \midrule
    \multicolumn{9}{c}{Our transformed CNNs}\\\midrule
    T-ResNet50-RS  & 224 & 38 M  & 447 &  17.6 & 81.0         & 48.0 & 18.7 & 42.9 \\
    T-ResNet101-RS & 224 & 66 M  & 334 &  25.1 & 82.4         & 52.9 & 27.7 & 47.8 \\
    T-ResNet152-RS & 320 & 89 M  & 128 &  65.8 & 83.7         & 54.5 & 39.8 & 50.6 \\
    T-ResNet200-RS & 320 & 96 M  & 105 &  80.2 & 84.0         & 57.0 & 41.2 & 51.1 \\
    T-ResNet270-RS & 320 & 133 M &  75 & 107.2 & 84.3         & 58.6 & 43.7 & 51.4 \\
    T-ResNet350-RS & 320 & 167 M &  61 & 130.5 & \textbf{84.5}& 59.2 & \textbf{44.8} & \textbf{53.8}\\
    \bottomrule
    \end{tabular}
    \caption{\textbf{Accuracy of our models on various benchmarks.} Throughput is the number of images processed per second on a V100 GPU at batch size 32. The ViT and ResNet results are reported in~\cite{bhojanapalli2021understanding}. For ImageNet-C, we keep a resolution of 224 at test time to avoid distorting the corruptions.
    }
    \label{tab:robustness}
\end{table}

\paragraph{Robustness gains}

%\ari{Given that the other papers are recent, do we want to use them as the motivating idea for this experiment or frame them as concurrent work which agree with our conclusions?}
Recent work~\cite{bhojanapalli2021understanding,mao2021rethinking} has shown that Transformer-based architectures are more robust to input perturbations than convolutional architectures. We therefore investigate whether our fine-tuning procedure brings robustness gains to the original CNNs. To do so, we consider three benchmarks. First, ImageNet-C~\cite{hendrycks2019robustness}, a dataset containing 15 sets of randomly generated corruptions, grouped into 4 categories: ‘noise’, ‘blur’, ‘weather’, and ‘digital’. Each corruption type has five levels
of severity, resulting in 75 distinct corruptions. Second, ImageNet-A~\cite{hendrycks2021nae}, a dataset containing naturally ``adversarial'' examples from ImageNet. Finally, we evaluate robustness to distribution shifts with ImageNet-R \cite{hendrycks2020many}, a dataset with various stylized “renditions” of
ImageNet images ranging from paintings to embroidery, which strongly modify the local image statistics. 

As shown in Tab.~\ref{tab:robustness} and illustrated in Fig.~\ref{fig:intro}, the T-ResNet-RS substantially outperforms the ResNet-RS on all three benchmarks. For example, our T-ResNet101-RS reaches similar or higher top-1 accuracy than the ResNet200-RS on each task, despite its lower top-1 accuracy on ImageNet-1k. This demonstrates that SA improves robustness more than it improves classification accuracy. 

To better understand where the benefits come from, we decompose the improvement of the T-ResNet50-RS over the various corruption severeties and categories of ImageNet-C in Fig.~\ref{fig:robustness}. We observe that improvement increases almost linearly with corruption severity. Although performance is higher in all corruption categories, there is a strong variability: the T-CNN shines particularly in tasks where the objects in the image are less sharp due to lack of contrast, bad weather or blurriness. We attribute this to the ability of SA to distinguish shapes in the image, as investigated in Sec~\ref{sec:dissection}.

\section{Dissecting the Transformed CNNs}
\label{sec:dissection}

In this section, we analyze various observables to understand how the representations of a T-ResNet270-RS evolve from those of the ResNet270-RS throughout training.

\begin{figure*}[htb]
    \centering
    \includegraphics[width=\linewidth]{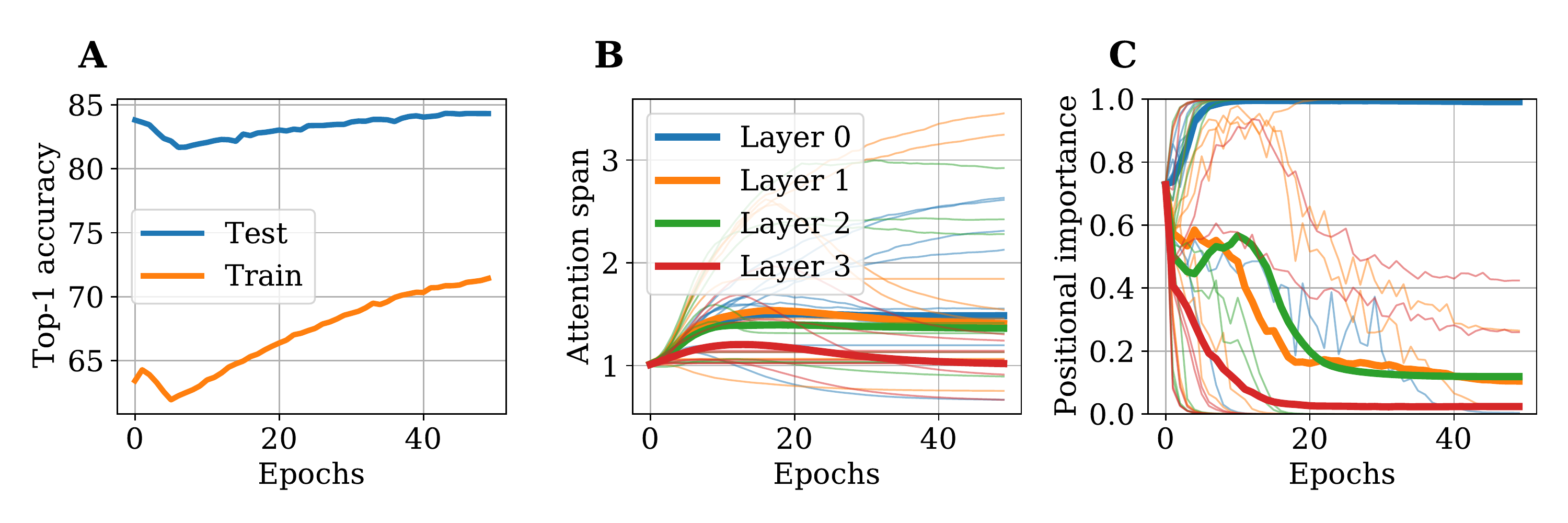}
    \caption{\textbf{The later layers effectively escape the convolutional configuration.} \textbf{A:} top-1 accuracy throughout the 50 epochs of fine-tuning of a T-ResNet270-RS. \textbf{B:} size of the receptive field of the various heads $h$ (thin lines), calculated as $\alpha_h^{-1}$ (see Eq.~\ref{eq:local-attention}). Thick lines represent the average over the heads. \textbf{C:} depicts how much attention the various heads $h$ (thin lines) pay to positional information, through the value of $\sigma(\lambda_h)$ (see Eq.~\ref{eq:gating-param}). Thick lines represent the average over the heads.}
    \label{fig:dynamics}
\end{figure*}

\vspace{.5cm}

\begin{figure}[htb]
    \centering
    \begin{subfigure}[b]{.24\textwidth}	   
     \includegraphics[width=\linewidth]{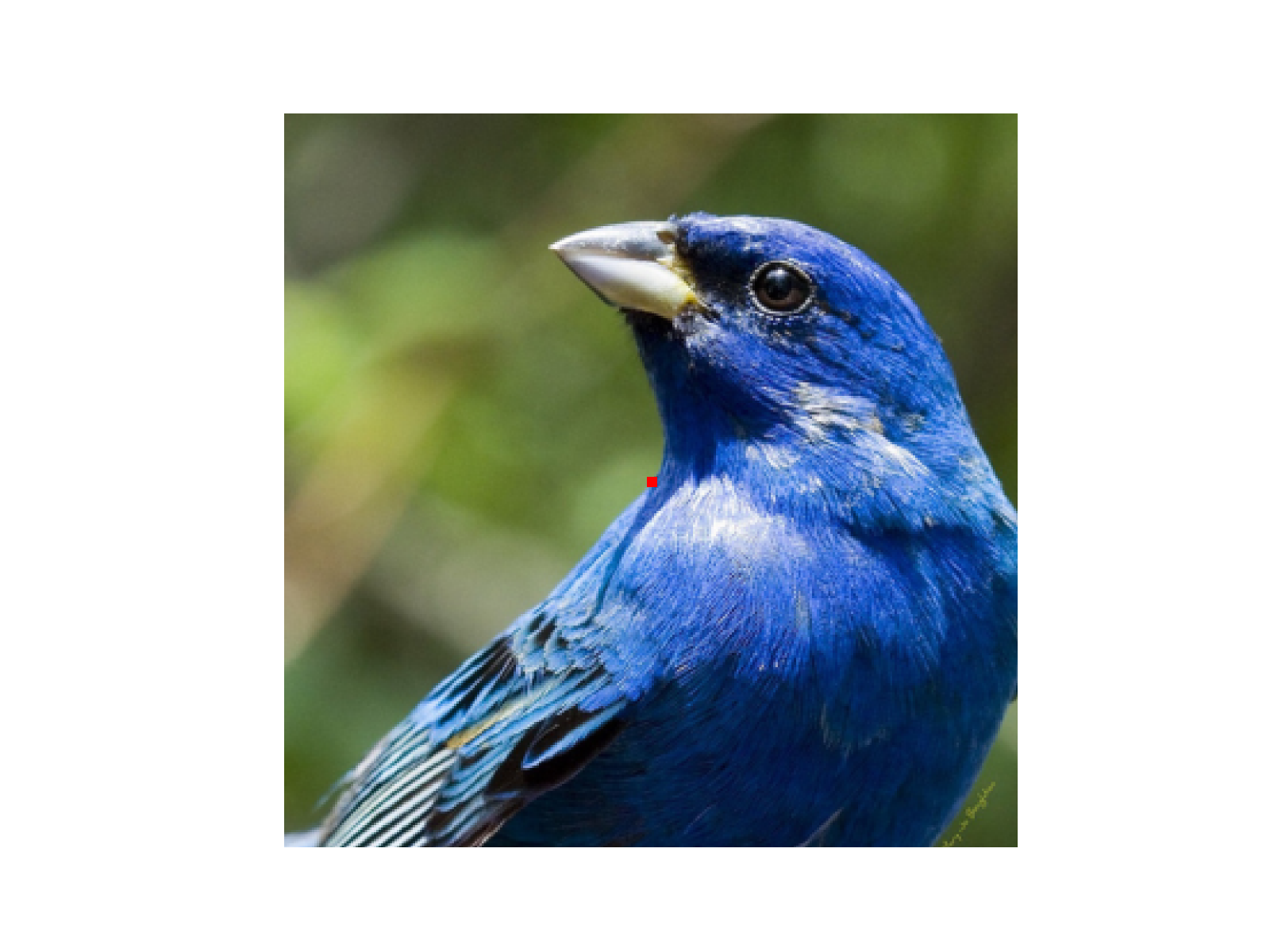}
     \vspace*{.1cm}
    \caption{Input image}
    \end{subfigure}
    \begin{subfigure}[b]{.75\textwidth}	   
    \includegraphics[width=\linewidth]{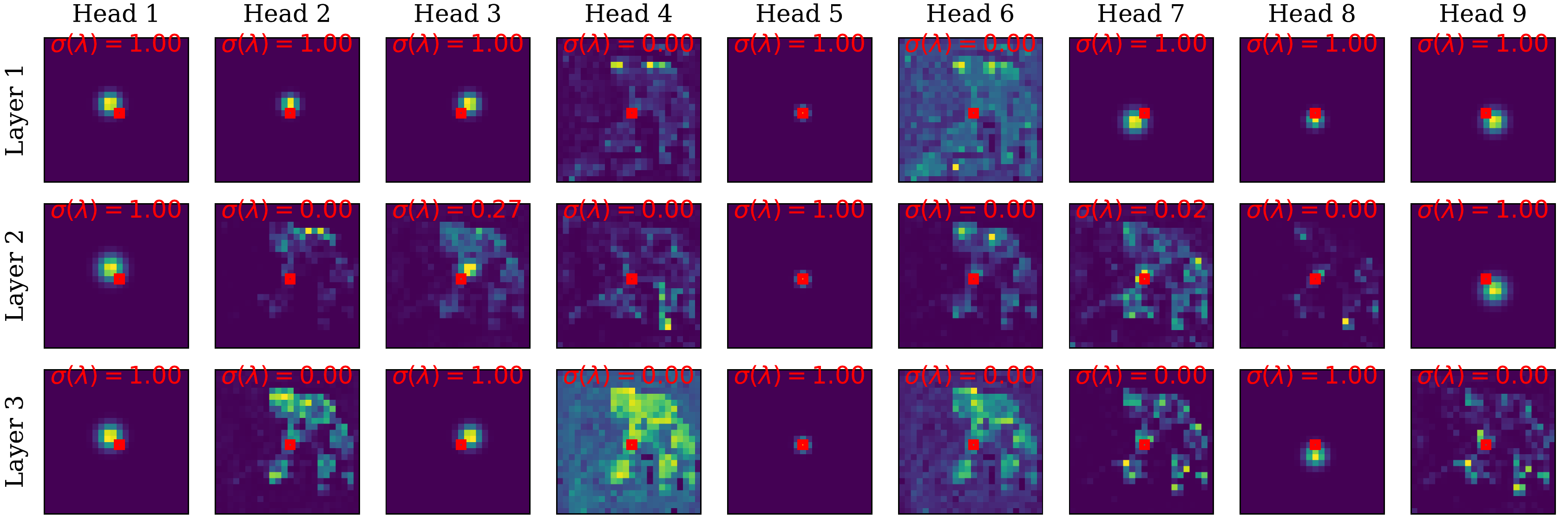}
    \caption{Attention maps}
    \end{subfigure}
    \caption{\textbf{GPSA layers combine local and global attention in a complementary way.} We depicted the attention maps of the four GPSA layers of the T-ResNet270-RS, obtained by feeding the image on the left through the convolutional backbone, then selecting a query pixel in the center of the image (red box). For each head $h$, we indicate the value of the gating parameter $\sigma(\lambda_h)$ in red (see Eq.~\ref{eq:gating-param}). In each layer, at least one of the heads learns to perform content-based attention ($\sigma(\lambda_h)=0$).}
    \label{fig:attention}
\end{figure}

\paragraph{Unlearn to better relearn} 
In Fig.~\ref{fig:dynamics}A, we display the train and test accuracy throughout training\footnote{The train accuracy is lower than the test accuracy due to the heavy data augmentation used during fine-tuning.}. The dynamics decompose into two distinct phases: accuracy dips down during the learning rate warmup phase (first 5 epochs of training), then increases back up as the learning rate is decayed. 

Interestingly, as shown in SM.~\ref{app:lr}, the depth of the dip depends on the learning rate. For too small learning rates, the dip is small, but the test accuracy increases too slowly after the dip; for too large learning rates, the test accuracy increases rapidly after the dip, but the dip is too deep to be compensated for. This suggests that the T-CNN needs to ``unlearn'' to some extent, a phenomenon reminiscent of the ``catapult'' mechanism of~\citet{lewkowycz2020large} which propels models out of sharp minima to land in wider minima.

\paragraph{Escaping the convolutional representation} 
In Fig.~\ref{fig:dynamics}B, we show the evolution of the ``attention span'' $1/\alpha_h$ (see Eq.~\ref{eq:local-init}), which reflects the size of the receptive field of attention head $h$. On average (thick lines), this quantity increases in the first three layers, showing that the attention span widens, but variability exists among different attention heads (thin lines): some broaden their receptive field, whereas others contract it.

In Fig.~\ref{fig:dynamics}C, we show the evolution of the gating parameters $\lambda^h$ of Eq.~\ref{eq:gating-param}, which reflect how much attention head $h$ pays to position versus content. Interestingly, the first layer stays strongly convolutional on average, as $\E_h \sigma(\lambda_h)$ rapidly becomes close to one (thick blue line). The other layers strongly escape locality, with most attention heads focusing on content information at the end of fine-tuning.

In Fig.~\ref{fig:attention}, we display the attention maps after fine-tuning. A clear divide appears between the ``convolutional'' attention heads, which remain close to their initialization, and the ``content-based'' attention heads, which learn more complex dependencies. Notice that the attention head initially focusing on the query pixel (head 5) stays convolutional in all layers. Throughout the layers, the shape of the central object is more and more clearly visible, as observed in~\cite{caron2021emerging}. This supports the hypothesis that robustness gains obtained for blurry corruptions (see Fig.~\ref{fig:robustness}) are partly due to the ability of the SA layers to isolate objects from the background. 

\section{When should one start learning the self-attention layers?}

% \begin{wraptable}{l}{.33\textwidth}
%     \centering
%     \begin{tabular}{c|c|c}
%     \toprule
%     Model & Res. & Top-1 \\
%     \midrule
%     ResNet50   & 224 & 79.04 \\
%     ResNet50   & 320 & 79.78 \\
%     T-ResNet50 & 224 & 79.88 \\
%     T-ResNet50 & 320 & \textbf{80.84} \\\bottomrule
%     \end{tabular}
%     \caption{\textbf{Fine-tuning yields significant performance gains.}
%     }
%     \label{tab:finetune-resnet50}
% \end{wraptable}

Previous sections have demonstrated the benefits of initializing T-CNNs from pre-trained CNNs, a very compelling procedure given the wide availability of pretrained models. But one may ask: how does this compare to training a hybrid model from scratch? More generally, given a computational budget, how long should the SA layers be trained compared to the convolutional backbone?

\paragraph{Transformed CNN versus hybrid models}
To answer the first question, we consider a ResNet-50 trained on ImageNet for 400 epochs. We use SGD with momentum 0.9 and a batch size of 1024, warming up the learning rate for 5 epochs before a cosine decay. To achieve a strong baseline, we use the same augmentation scheme as in~\cite{touvron2020training} for the DeiT. Results are reported in Tab.~\ref{tab:tw}. In this modern training setting, the vanilla ResNet50 reaches a solid performance of 79.04\% on ImageNet, well above the 77\% usually reported in litterature. 

\begin{table}%{l}{.5\textwidth}
    \centering
    \begin{tabular}{c|c|c|c|c}
    \toprule
    \textbf{Name} & $t_1$ & $t_2$ & Train time & Top-1 \\\midrule%& Reinit acc\\\midrule
    Vanilla CNN   & 400 & 0   & 2.0k mn & 79.04   \\
    Vanilla CNN$\uparrow$320 & 450 & 0  & 2.4k mn & \textbf{79.78}\\\midrule
    T-CNN & 400 & 50  & 2.3k mn & 79.88   \\
    T-CNN$\uparrow$320  & 400 & 50  & 2.7k mn & \textbf{80.84}   \\\midrule
    Vanilla hybrid & 0  & 400 & 2.8k mn & 79.95   \\%\midrule
    T-CNN$^\star$ & 100 & 300 & 2.6k mn & \textbf{80.44}   \\
    T-CNN$^\star$ & 200 & 200 & 2.4k mn & 80.28   \\
    T-CNN$^\star$ & 300 & 100 & 2.2k mn & 79.28   \\
    %400 & 50  & 2.3k mn & 79.88 \\        
    \bottomrule
    \end{tabular}
    \caption{\textbf{The benefit of late reparametrization.}  We report the top-1 accuracy of a ResNet-50 on ImageNet reparameterized at various times $t_1$ during training. $\uparrow$320 stands for fine-tuning at resolution 320. The models with a $\star$ keep the same optimizer after reparametrization, in contrast with the usual T-CNNs.}
    \label{tab:tw}
%     }
\end{table}

The T-CNN obtained by fine-tuning the ResNet for 50 epochs at same resolution obtains a top-1 accuracy of 79.88\%, with a 15\% increase in training time, and 80.84 as resolution 320, with a 35\% increase in training time. In comparison, the hybrid model trained for 400 epochs in the same setting only reaches 79.95\%, in spite of a 40\% increase in training time. Hence, fine-tuning yields better results than training the hybrid model from scratch.

\paragraph{What is the best time to reparametrize?}

We now study a scenario between the two extreme cases: what happens if we reparametrize halfway through training? To investigate this question in a systematic way, we train the ResNet50 for $t_1$ epochs, then reparametrize and resume training for another $t_2$ epochs, ensuring that $t_1+t_2=400$ in all cases. Hence, $t_1=400$, amounts to the vanilla ResNet50, whereas $t_1=0$ corresponds to the hybrid model trained from scratch. To study how final performance depends on $t_1$ in a fair setting, we keep the same optimizer and learning rate after the reparametrization, in contrast with the fine-tuning procedure which uses fresh optimizer.

Results are presented in Tab.~\ref{tab:tw}. Interestingly, the final performance evolves non-monotonically, reaching a maximum of $80.44$ for $t_1=100$, then decreasing back down as the SA layers have less and less time to learn. This non-monotonicity is remarkably similar to that observed in~\cite{d2019finding}, where reparameterizing a CNN as a FCN in the early stages of training enables the FCN to outperform the CNN. Crucially, this result suggests that reparametrizing during training not only saves time, but also helps the T-CNN find better solutions.

% \begin{figure}
%     \centering
%     \includegraphics[width=\linewidth]{figs/acc_vs_time_CIFAR100.pdf}
% \caption{\textbf{There is an optimal time to train the SA.}  We report the performances obtained by a ResNet-50 trained from scratch on CIFAR100, reparameterized at various times $t_w$ during training. }
%     \label{fig:cifar}
% \end{figure}

%\begin{wraptable}{r}{5.5cm}
% \begin{table}
%     \centering
%     \begin{tabular}{c|c|c|c}
%     \toprule
%     $t_w$ (epochs) & Train time (minutes) & Top-1 (load filters) & Top-1 (reinit filters)\\\midrule
%     400 (CNN)  & 2.0k & 79.04          & 79.04\\
%     300        & 2.2k & 79.28          & 79.76\\
%     200        & 2.4k & 80.28          & 80.22\\
%     100        & 2.6k & \textbf{80.44} & \textbf{80.38}\\
%     0  (hybrid)& 2.8k & 80.14          & 79.95\\
%     % 300  & 3.1k & 79.65      \\
%     % 200  & 4.3k & \textbf{80.36}\\
%     % 100  & 5.5k & 79.80      \\
%     % 0    & 6.6k & 79.71      \\
%     \bottomrule
%     \end{tabular}
%     \caption{\textbf{There is an optimal time to unleash the SA.}  We report the performances obtained by a ResNet-50 trained from scratch on ImageNet-1k, reparameterized at various times $t_w$ during training.
%     }
%     \label{tab:hybrid}
% \end{table}

\section*{Discussion}

In this work, we showed that complex building blocks such as self-attention layers need not be trained from start. Instead, one can save in compute time while gaining in performance and robustness by initializing them from pre-trained convolutional layers. At a time where energy savings and robustness are key stakes, we believe this finding is important. 

On the practical side, our fine-tuning method offers an interesting new direction for practitioners. One clear limitation of our method is the prohibitive cost of reparametrizing the early stages of CNNs. This cost could however be alleviated by using linear attention methods~\cite{wang2020self}, an important direction for future work. Note also that while our T-CNNs significantly improve the robustness of CNNs, they do not systematically reach the performance of end-to-end Transformers such as the DeiT (for example on ImageNet-C, see Fig.~\ref{fig:intro}). Bridging this gap is an important next step for hybrid models.

On the theoretical side, our results spark several interesting questions. First, why is it better to reparametrize at intermediate times? One natural hypothesis, which will be explored in future work, is that SA layers benefit from capturing meaningful dependencies between the features learnt by the CNN, rather than the random correlations which exist at initialization. Second, why are the representations learnt by the SA layers more robust? By inspecting the attention maps and the most improved corruption categories of ImageNet-C, we hypothesized that SA helps isolating objects from the background, but a more thorough analysis is yet to come. 
\paragraph{Acknowledgements}
We thank Matthew Leavitt, Hugo Touvron, Hervé Jégou and Francisco Massa for helpful discussions. SD and GB acknowledge funding from the French government under management of Agence Nationale de la Recherche as part of the “Investissements d’avenir” program, reference ANR-19-P3IA-0001 (PRAIRIE 3IA Institute). 

\printbibliography

\appendix

\numberwithin{equation}{section}% \renewcommand{\theequation}{S.\arabic{equation}}

\clearpage

\appendix
\setlength\intextsep{10pt}

\section{Changing of learning rate}
\label{app:lr}

\begin{figure}[h]
    \centering
    \includegraphics[width=.45\textwidth]{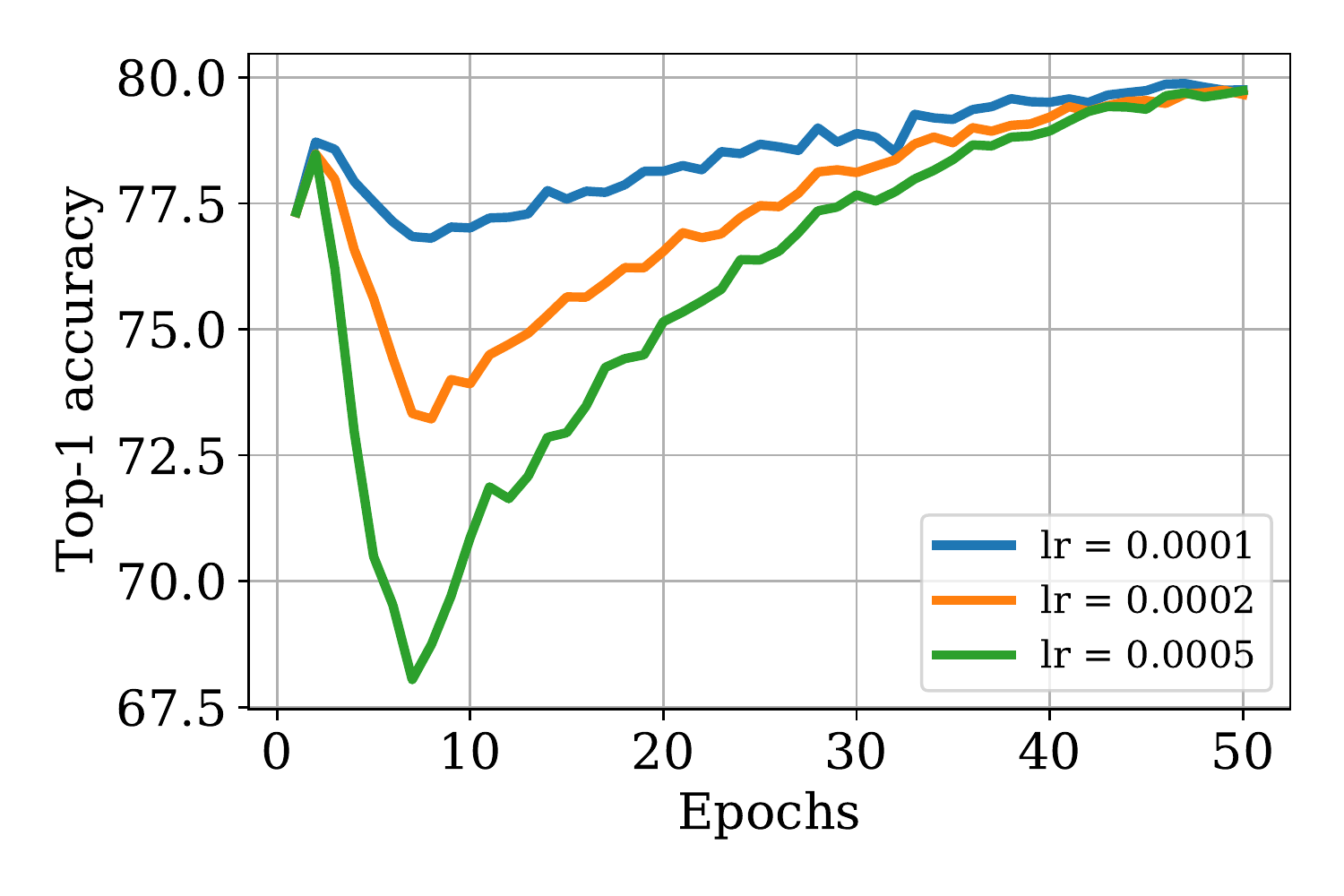}
    \caption{\textbf{The larger the learning rate, the lower the test accuracy dips, but the faster it climbs back up.} We show the dynamics of the ResNet50, fine-tuned for 50 epochs at resolution 224, for three different values of the maximal learning rate.}
    \label{fig:lr}
\end{figure}

As shown in Fig.~\ref{fig:dynamics} of the main text, the learning dynamics decompose into two phases: the learning rate warmup phase, where the test loss drops, then the learning rate decay phase, where the test loss increases again. This could lead one to think that the maximal learning rate is too high, and the dip could be avoided by choosing a lower learning rate. Yet this is not the case, as shown in Fig.~\ref{fig:lr}. Reducing the maximal learning rate indeed reduces the dip, but it also slows down the increase in the second phase of learning. This confirms that the model needs to ``unlearn'' the right amount to find better solutions.

\section{Changing the test resolution}
\label{app:resolution}

One advantage of the GPSA layers introduced by~\cite{d2021convit} is how easily they adapt to different image resolutions. Indeed, the positional embeddings they use are fixed rather than learnt. They simply consist in 3 values for each pair of pixels: their euclidean distance $\Vert \bs \delta\Vert$, as well as their coordinate distance $\bs \delta_1, \bs \delta_2$ (see Eq.~\ref{eq:local-init}). Our implementation automatically adjusts these embeddings to the input image, allowing us to change the test resolution seamlessly.

\begin{figure}
    \centering
    \includegraphics[width=.6\linewidth]{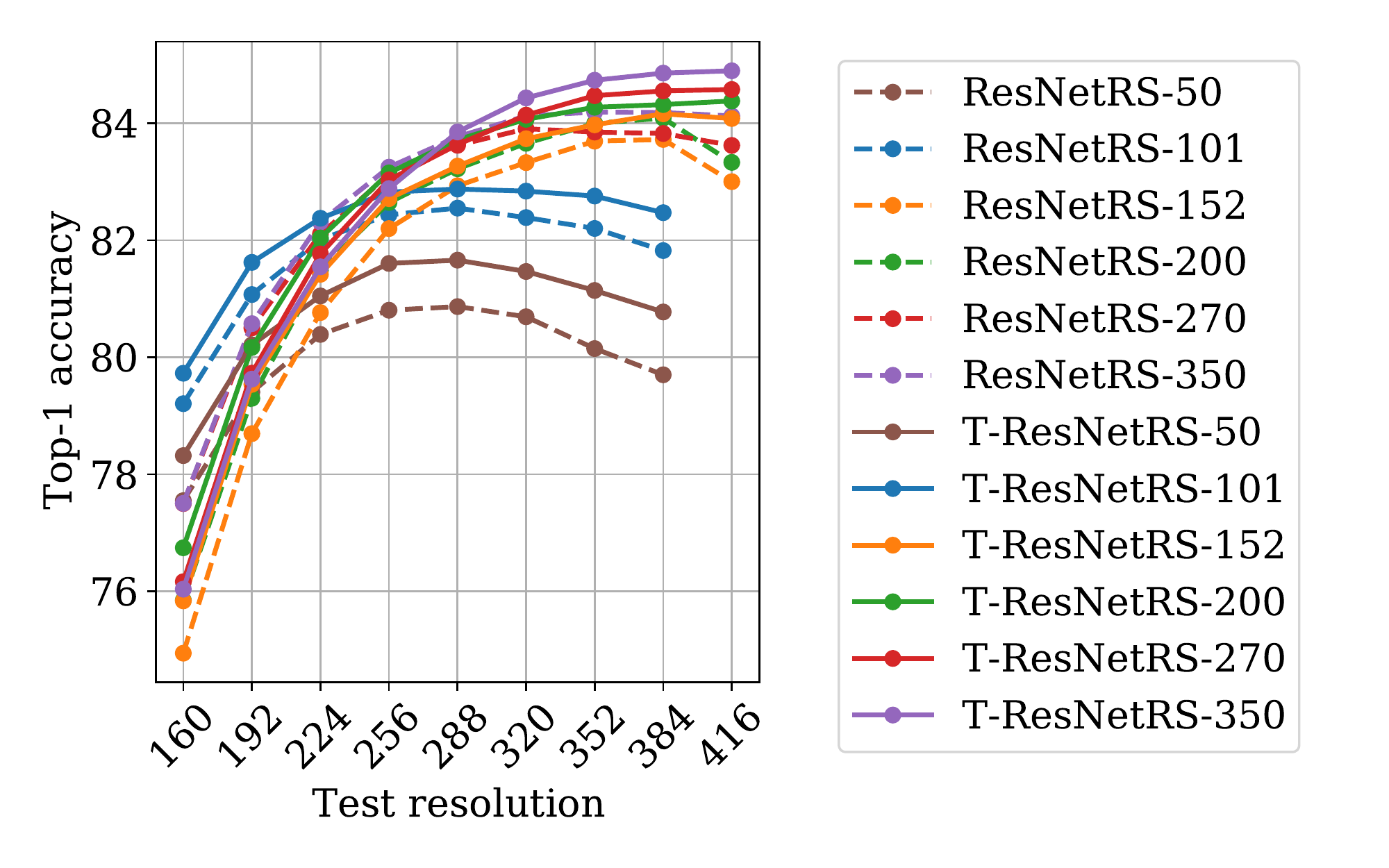}
    \caption{\textbf{Performance at different test-time resolutions, for the finetuned models with and without SA.} The ResNet50-RS and ResNet101-RS models are finetuned at resolution 224, and all other models are finetuned at resolution 320.}
    \label{fig:resolution}
\end{figure}

In Fig.~\ref{fig:resolution}, we show how the top-1 accuracies of our T-ResNet-RS models compares to those of the ResNet-RS models finetuned at same resolution but without SA. At test resolution 416, our T-ResNetRS-350 reaches an impressive top-1 accuracy of 84.9\%, beyond those of the best EfficientNets and BotNets~\cite{srinivas2021bottleneck}.

\section{Changing the number of epochs}
\label{app:epochs}

In Tab.~\ref{tab:epochs}, we show how the top-1 accuracy of the T-ResNet-RS model changes with the number of fine-tuning epochs. As expected, performance increases significantly as we fine-tune for longer, yet we chose to set a maximum of 50 fine-tuning epochs to keep the computational cost of fine-tuning well below that of the original training.

\begin{table}[h]
    \centering
    \begin{tabular}{c|c|c}
    \toprule
        \textbf{Model}   & Epochs & Top-1 acc \\\midrule
        ResNet50-RS & 0               & 79.91 \\
        T-ResNet50-RS & 10              & 80.11 \\
        T-ResNet50-RS & 20              & 80.51 \\
        T-ResNet50-RS & 50              & \textbf{81.02} \\\midrule 
        ResNet101-RS & 0              & 81.70 \\
        T-ResNet101-RS & 10             & 81.54 \\
        T-ResNet101-RS & 20             & 81.90 \\
        T-ResNet101-RS & 50             & \textbf{82.39} \\
    \bottomrule
    \end{tabular}
    \caption{\textbf{Longer fine-tuning increases final performance.} We report the top-1 accuracies of our models on ImageNet-1k at resolution 224.}
    \label{tab:epochs}
\end{table}

\section{Changing the architecture}
\label{app:resnetd}

Our framework, which builds on the timm package, makes changing the original CNN architecture very easy. We applied our fine-tuning procedure to the ResNet-D models~\cite{he2019bag} with the exact same hyperparameters, and observed substantial performance gains, similar to the ones obtained for ResNet-RS, see Tab.~\ref{tab:finetune-resnetd}. This suggests the wide applicability of our method.

\begin{table}[h]
    \centering
    \begin{tabular}{c|c|c|c|c|c}
    \toprule
    \textbf{Model} & Original res. & Original acc. & Fine-tune res. & Fine-tune acc. & Gain\\
    \midrule
    %T-ResNet26-D   & 224 & 76.7 & 320 & 78.1  & +1.4\\
    T-ResNet50-D   & 224 & 80.6 & 320 & 81.6  & +1.0\\
    T-ResNet101-D  & 320 & 82.3 & 384 & 83.1  & +0.8\\
    T-ResNet152-D  & 320 & 83.1 & 384 & 83.8  & +0.7\\
    T-ResNet200-D  & 320 & 83.2 & 384 & \textbf{83.9}  & +0.7\\
    \midrule
    T-ResNet50-RS  & 160 & 78.8 & 224 & 81.0  & +2.8\\
    T-ResNet101-RS & 192 & 81.2 & 224 & 82.4  & +1.2\\
    T-ResNet152-RS & 256 & 83.0 & 320 & 83.7  & +0.7\\
    T-ResNet200-RS & 256 & 83.4 & 320 & \textbf{84.0}  & +0.6\\
    \bottomrule
    \end{tabular}
    \caption{\textbf{Comparing the performance gains of the ResNet-RS and ResNet-D architectures.} Top-1 accuracy is measured on ImageNet-1k validation set. The pre-trained models are all taken from the timm library~\cite{rw2019timm}.
    }
    \label{tab:finetune-resnetd}
\end{table}

\clearpage
\section{More attention maps}

\begin{figure}[htb]
    \centering
    \begin{subfigure}[b]{.24\textwidth}	   
     \includegraphics[width=\linewidth]{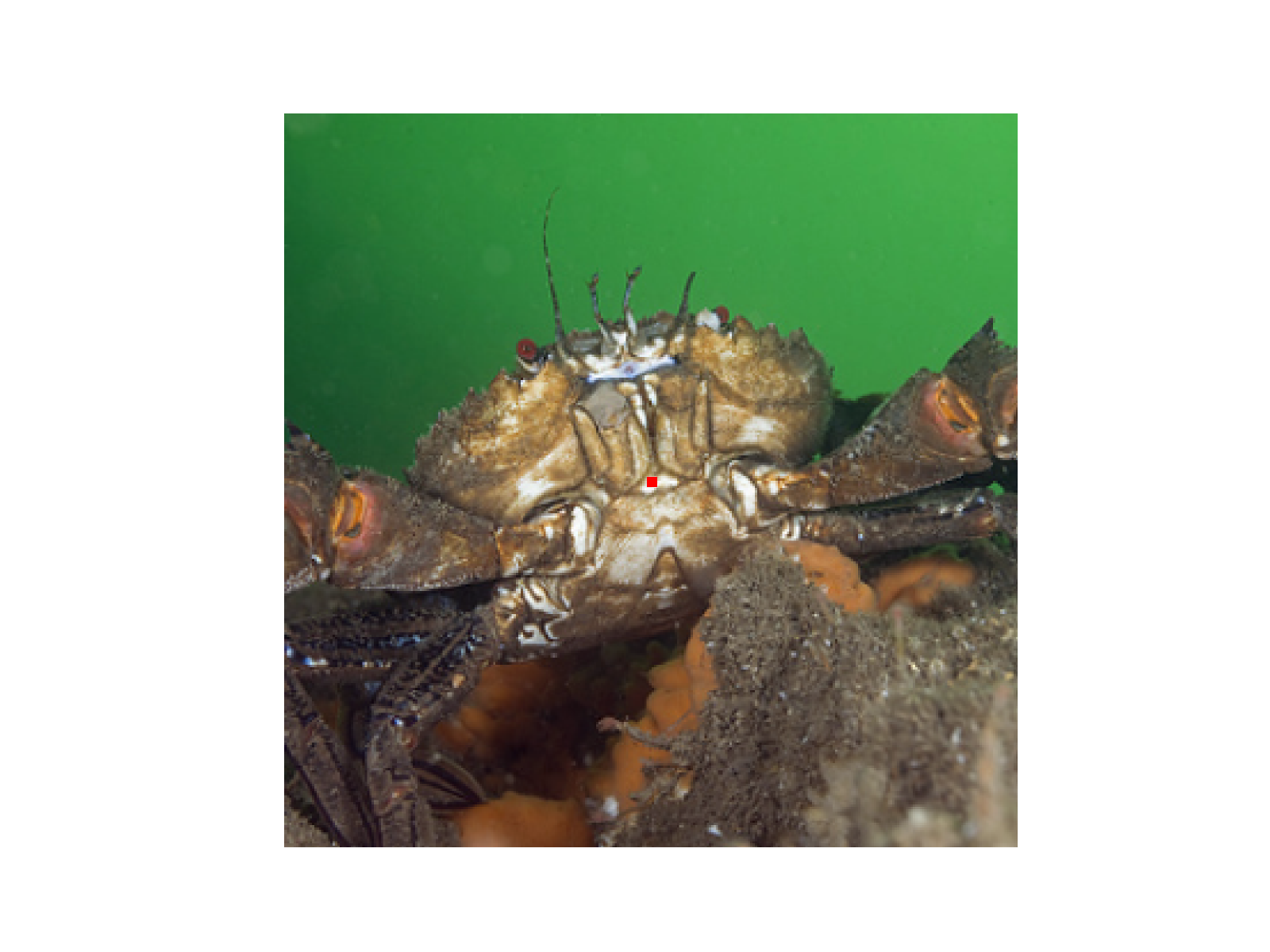}
     \vspace*{.6cm}
    \end{subfigure}
    \begin{subfigure}[b]{.75\textwidth}	   
    \includegraphics[width=\linewidth]{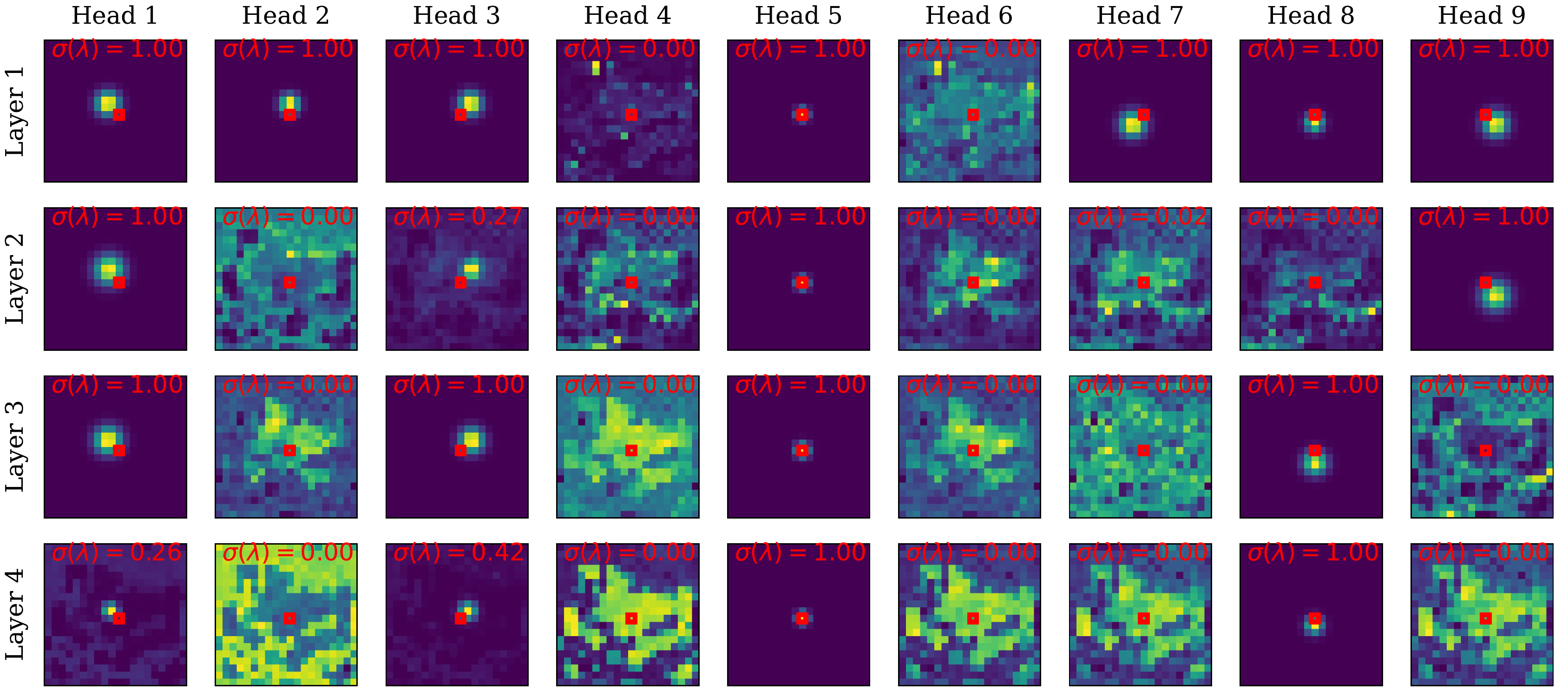}
    \caption{Attention maps}
    \end{subfigure}
    \begin{subfigure}[b]{.24\textwidth}	   
     \includegraphics[width=\linewidth]{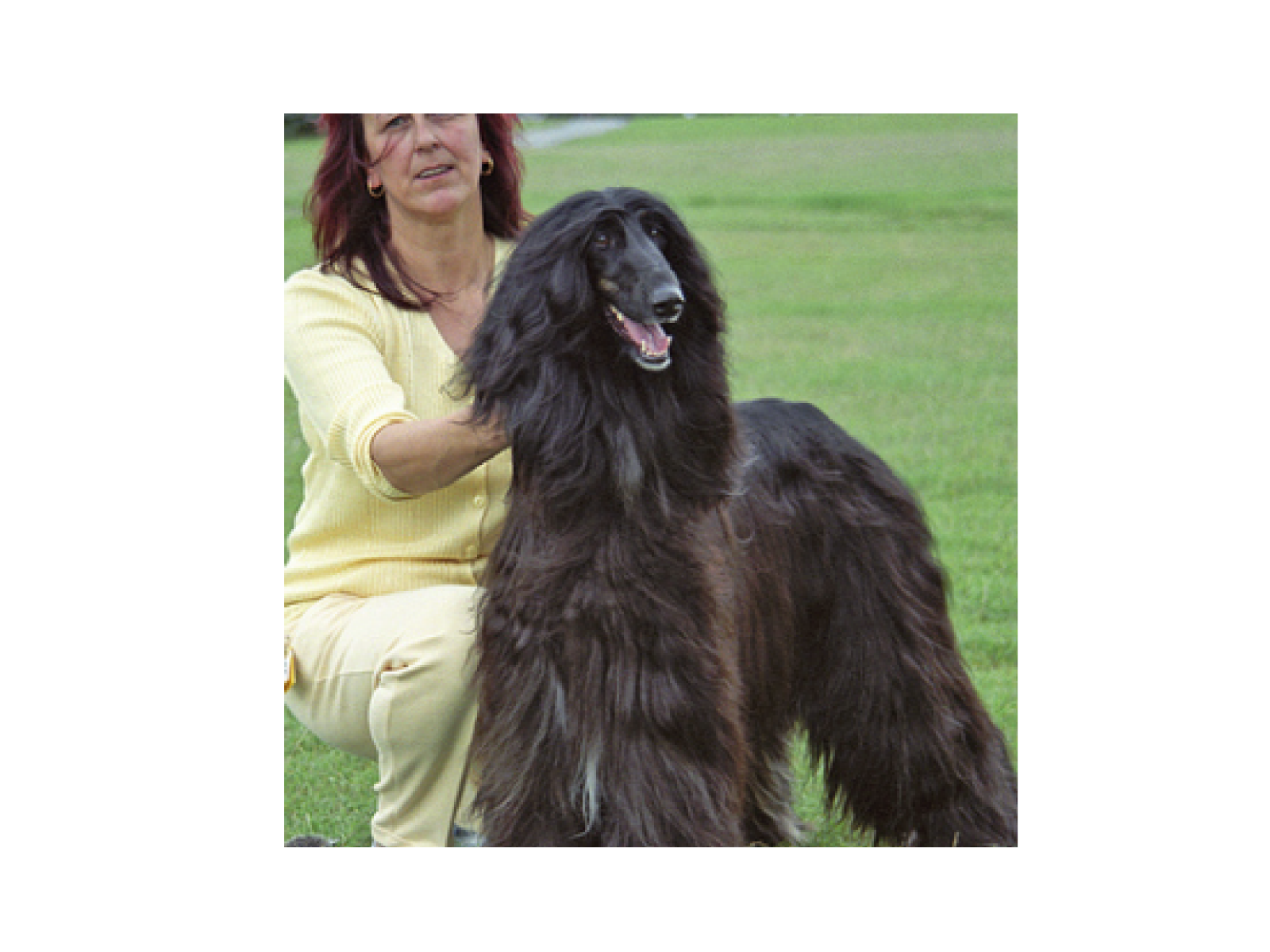}
     \vspace*{.6cm}
    \end{subfigure}
    \begin{subfigure}[b]{.75\textwidth}	   
    \includegraphics[width=\linewidth]{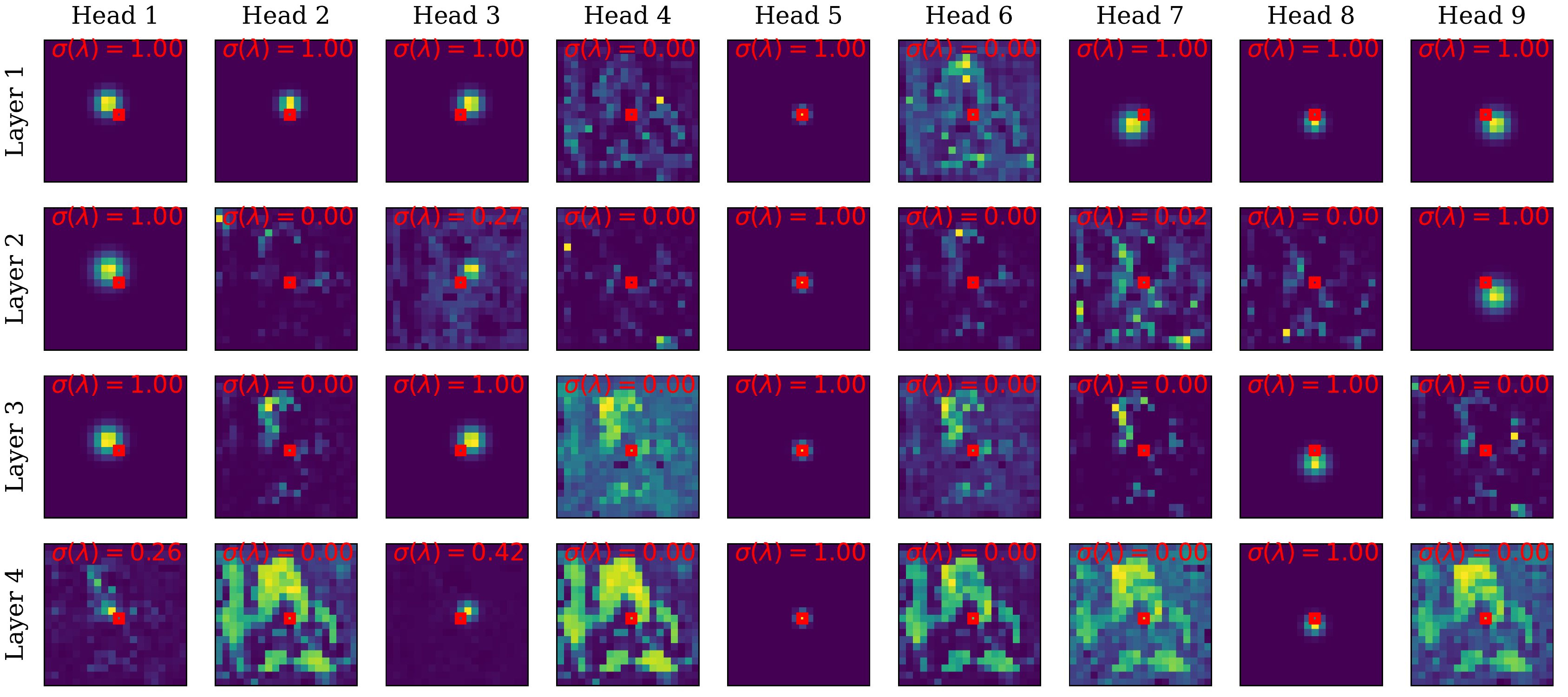}
    \caption{Attention maps}
    \end{subfigure}    
    \caption{\textbf{GPSA layers combine local and global attention in a complementary way.} We depicted the attention maps of the four GPSA layers of the T-ResNet270-RS, obtained by feeding the image on the left through the convolutional backbone, then selecting a query pixel in the center of the image (red box). For each head $h$, we indicate the value of the gating parameter $\sigma(\lambda_h)$ in red (see Eq.~\ref{eq:gating-param}).  ($\sigma(\lambda_h)=0$).}
    \label{fig:more-attention}
\end{figure}

\clearpage
\end{document}